# Feedback modalities for a table setting robot assistant for elder care

Noa Markfeld, Samuel Olatunji, Dana Gutman, Shay Givati,
Vardit Sarne-Fleischmann, Yael Edan
Ben-Gurion University of the Negev, Beer-Sheva, Israel

**Abstract.** Older adults' interaction with robots requires effective feedback to keep them aware of the state of the interaction for optimum interaction quality. This study examines the effect of different feedback modalities in a table setting robot assistant for elder care. Two different feedback modalities (visual and auditory) and their combination were evaluated for three complexity levels. The visual feedback included the use of LEDs and a GUI screen. The auditory feedback included alerts (beeps) and verbal commands. The results revealed that the quality of interaction was influenced mainly by the feedback modality, and complexity had less influence. The verbal feedback was significantly preferable and increased the participants' involvement during the experiment. The combination of LED lights and verbal commands increased participants' understanding contributing to the quality of interaction.

**Keywords:** human-robot interaction, feedback modalities, collaborative robot, assistive robot, older adults.

## 1      Introduction

The world's elderly population is rapidly growing due to the increase in life expectancy [1]. However, the population of caregivers does not increase at a similar rate, leading to an increased need in developing solutions that will assist the older adults. One solution is the use of Socially Assistive Robots (SARs) to meet the needs of these older adults [2]. The older person has perceptual abilities distinct from those of the younger population particularly evident in processing information [3]. Moreover, what makes the older adult population such a unique group is that declines in abilities related to aging are not homogeneous [4]. Therefore, the correct choice of interfaces between the assisting environment and the user is of high importance [2]. Older adults' interaction with robots requires effective feedback to keep them aware of the state of the interaction for optimum interaction quality [3].

Feedback from the robot can help humans to evaluate the robot's internal state and its overall goals [5]. Existing studies reveal that the information presented to the user significantly influences his / her comprehension of the robot's behavior, performance and limitations of the robot [6], influencing interaction quality [4]. Robots can provide information to the human by visual feedback (using a screen or lights) [8], verbal feedback (warning noises or verbal commands) [7], and tactile devices [7]. Combinations of these modalities, multimodal feedback, may enhance user interactions [9] and can increase the quantity and quality of information conveyed [10]. Creating the most appropriate type of feedback is a main challenge in human-robot interaction [6].

This study evaluates various feedback modalities that the robot provides to the person when performing a joint task, focusing on two main feedback types–visual and auditory. These feedback types and their combination are evaluated for different complexity levels. The overall goal is to ensure high interaction quality between the older adult and the robot in accomplishing the desired task while increasing the older adult's satisfaction along the collaboration.

## 2      Methods

A human-robot collaborative system was developed to examine the effect of different types of feedback at different complexity levels on the interaction quality. A KUKA LBR iiwa 14 R820 7 degrees of freedom robotic arm equipped with a pneumatic gripper was programmed in a table-setting task performed jointly by an older adult and the robot. The tasks were programmed using the Python programming language and executed on the ROS platform. Two main types of feedback were examined - visual and auditory, and their combination. When providing visual feedback, both a graphical user interface (GUI) and LED lights were used. The GUI was presented on a PC screen (Figure 1a), which was located on a desk to the left of the user, whereas the LED lights were embedded in the robot and were connected to the system using a Raspberry Pi computer (Figure 1b). Audial feedback was transmitted to the user through a speaker system connected to the main computer and included using of beep alert and verbal commands.

Each type of feedback was evaluated for three complexity levels. The simple level provided feedback through the use of non-continuous alerts, using flashing lights and beeps. The intermediate level conveyed more information by using screen and verbal commands (for ex. "The task is starting") and the complicated level combined the previous two levels together. 21 older adults aged 70-86 participated in the study (mean 74, std 4.12) which was designed as a between-within experiment with the type and complexity of feedback defined as the independent variables. Each participant experienced one type of feedback while performing the task at three levels of complexity in a random order.

The dependent variable was the quality of the interaction which consisted of trust, engagement, understanding and comfortability measures. The measures were selected based on the relevance of these measures to the older adult population found in previous studies as detailed below. These variables were assessed subjectively through questionnaires used 5-point Likert scales with 5 representing "*Strongly agree*" and 1 representing "*Strongly disagree*", and objectively through recorded videos which were manually analyzed. The *trust* measure shows the level of reliance on the robot to enjoy successful interaction [12], evaluated by analysing the participant's sitting position and proximity to the robot (three positions were pre-defined and offered to participants before each session). The *engagement* measure describes the amount of time there was eye contact between humans and the robot implying the relationship between the older adult and the robot [13]. This measure is very significant for the elderly population who may lose attention and therefore must be kept consistently in the loop and as active as possible in the interaction [14]. *Understanding* is required for the robot and human to be able to successfully interact with each other [15]. Its important to assess the degree of understanding that the user has in the interaction [16] in order to ensure adequate situation awareness [17]. This indicator examines whether the feedback was clear to the user evaluated by the amount of clarifications the person requested. The *comfortability* measure influences the level of satisfaction the user has while interacting with the system [18] and how much feedback was provided was convenient and accessible to the user. This measure was evaluated by the difference in the user's heart rate during the session and by the amount of physical gestures the user made.

Initially, participants completed three pre-test questionnaires: 1. a questionnaire which included demographic information, 2. a subset of questions from the Technology Adoption Propensity (TAP) index to assess their level of experience with technology [19] and 3. a sub-set from the Negative Attitude toward Robots Scale (NARS) to assess their level of anxiety towards robots [20]. The TAP and NARS questionnaires used a 5-point Likert scale.

A two tailed General Linear Mixed Model (GLMM) analysis was performed to evaluate for a positive or negative effect of the independent variables.

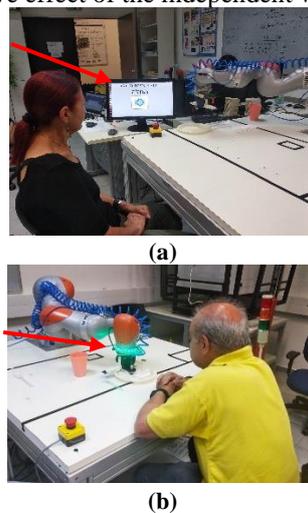

**Fig. 1.** GUI (a) and LED (b) feedback with arrow pointing on the feedback.

## 3 Results

Most of the participants (78%) were comfortable interacting with a robot. The results revealed that the quality of interaction, as measured via trust, engagement, understanding and comfortability of the interaction was influenced mainly by the type of feedback ($p = 0.05$), and complexity had less influence ($p = 0.24$). For each type of feedback, the participants indicated a specific preference for the different levels of complexity as detailed below.

### 3.1 Audial feedback

The preferred feedback for the audial type was the verbal feedback (implemented at the medium complexity level), a result which was reflected in all measures. Using this feedback increased the involvement of the participants during the experiment. i.e. the number of subjects' comments was higher (p = 0.024). Also, 86% of participants indicated that the verbal feedback helped them to understand the robot best. 13% preferred the combination of beep and verbal commands (the high complexity level), and only 1% preferred the use of beeping (the low complexity level). The comfortability measure showed similar results. The most comfortable feedback was the one that contained the verbal commands (med = 2.4). The two levels that used the beeps were inconvenient to the users (med = 1.31). Also, heart rate during verbal commands was low (mean = 100.89) whereas beep feedback resulted in a higher rate (mean = 112.28). Moreover, there was a large difference in the participants' sense of trust in the robot between complexity levels (p=0.049), with verbal feedback showing a higher trust (med = 3.43) vs. beep only (med = 2.43).

### 3.2 Visual feedback

This feedback type was also consistent with all measures. The preferred feedback was the use of LED lights as 96% of the participants were focused on the robot during the task, and did not notice the information received from the screen. The simplest level of complexity involved in using LED lights resulted in the highest understanding (med = 3) compared to the two more complex levels that contained information displayed on a screen (med = 1.6). This preference was also noticeable in the comfortability measure. When using LED lights only, the overall sense of comfort was high (med = 2.3) and the heart rate measure was the lowest (mean = 98.96) whilst using the screen resulted in a lower sense of comfortability (med = 1.5), and a higher heart rate (mean = 115.07). When using LED lights, the lowest complexity level achieved the highest trust level (med = 3.03). This is probably due to the fact that using lights is similar to using other familiar devices.

### 3.3 Multimodal feedback

The multimodal feedback type provided the best understanding at all complexity levels (med = 3.8, p = 0.017). The levels containing verbal commands at the higher complexity levels, increased the understanding of the participants. The combination that contributed most to understanding was the combination of verbal commands and LED lights. The multimodal feedback contributed to the user's comfort and at all levels of complexity, mean heart rate was low (mean = 98). In both the comfortability measure and the trust measure, the most convenient (med = 3.1) and most reliable (med = 3.84) combination was the combination of LED lights and verbal commands (med = 3.1). A statistically significant result (p = 0.05) was obtained, showing a difference between the feedback types according to subjects' pleasure. Using multimodal feedback type showed greater pleasure, participants felt more natural with this type of feedback (med = 2.78). In addition, in-depth observation shows that the feedback that provided the greatest pleasure was the integration of LEDs into verbal commands (med = 3.14).

## 4   Conclusions

Verbal commands were found as the preferred feedback. This feedback significantly increased participants' involvement in the task, and it was evident that its use encouraged communication between the participant and the robot. The LED lights significantly contributed to understanding and also had a positive effect on the quality of interaction, probably due to their familiarity among older people derived from other devices they use. The combination of several components of feedback of the same modality did not contribute to the quality of interaction and even hampered the attention of the participants. However, using an intercensal combination of feedback increased interaction quality and succeeded in providing information to older participants in a way that suited their non-homogeneous abilities.

**Acknowledgements**

This research was supported by the EU funded Innovative Training Network (ITN) in the Marie Skłodowska-Curie People Programme (Horizon2020): SOCRATES (Social Cognitive Robotics in a European Society training research network), grant agreement number 721619. Partial support was provided by Ben-Gurion University of the Negev Agricultural, Biological and Cognitive Robotics Initiative, and the Rabbi W. Gunther Plaut Chair in Manufacturing Engineering.